\documentclass[journal]{vgtc}                     

\vgtccategory{Research}

\vgtcpapertype{system}

\title{InFiConD: Interactive No-code Fine-tuning with Concept-based Knowledge Distillation}

\author{%
  \authororcid{Jinbin Huang},
  Wenbin He, Liang Gou, Liu Ren and 
  Chris Bryan
}

\authorfooter{
  \item
  	Jinbin Huang and Chris Bryan are with Arizona State Uiversity.
  	E-mail: \{jhuan196, chris.bryan\}@asu.edu.
  \item
  	Wenbin He, Liang Gou and Liu Ren are with Bosch USA. E-mail:\{wenbin.he2, liu.ren\}@us.bosch.com.
}

\newcommand{\system}{\textsc{InFiCond}}

\newcommand{\concept}[1]{\texttt{#1}}
\newcommand{\class}[1]{\textit{#1}}
\newcommand{\user}[1]{Sam}

\abstract{
The emergence of large-scale pretrained models has heightened their application in various downstream tasks, yet deployment is a challenge in environments with limited computational resources. Knowledge distillation has emerged as a solution in such scenarios, whereby knowledge from large teacher models is transferred into smaller student' models, but this is a non-trivial process that traditionally requires technical expertise in AI/ML. To address these challenges, this paper presents \system{}, a novel framework that leverages visual concepts to implement the knowledge distillation process and enable subsequent no-code fine-tuning of student models. We develop a novel knowledge distillation pipeline based on extracting text-aligned visual concepts from a concept corpus using multimodal models, and construct highly interpretable linear student models based on visual concepts that mimic a teacher model in a response-based manner. \system{}’s interface allows users to interactively fine-tune the student model by manipulating concept influences directly in the user interface. We validate \system{} via a robust usage scenario and user study. Our findings indicate that \system{}’s human-in-the-loop and visualization-driven approach enables users to effectively create and analyze student models, understand how knowledge is transferred, and efficiently perform fine-tuning operations. We discuss how this work highlights the potential of interactive and visual methods in making knowledge distillation and subsequent no-code fine-tuning more accessible and adaptable to a wider range of users with domain-specific demands.
}
\keywords{Knowledge distillation, visual concepts, vision-language models, visual analytics.}

\teaser{
  \centering
  \includegraphics[width=\linewidth]{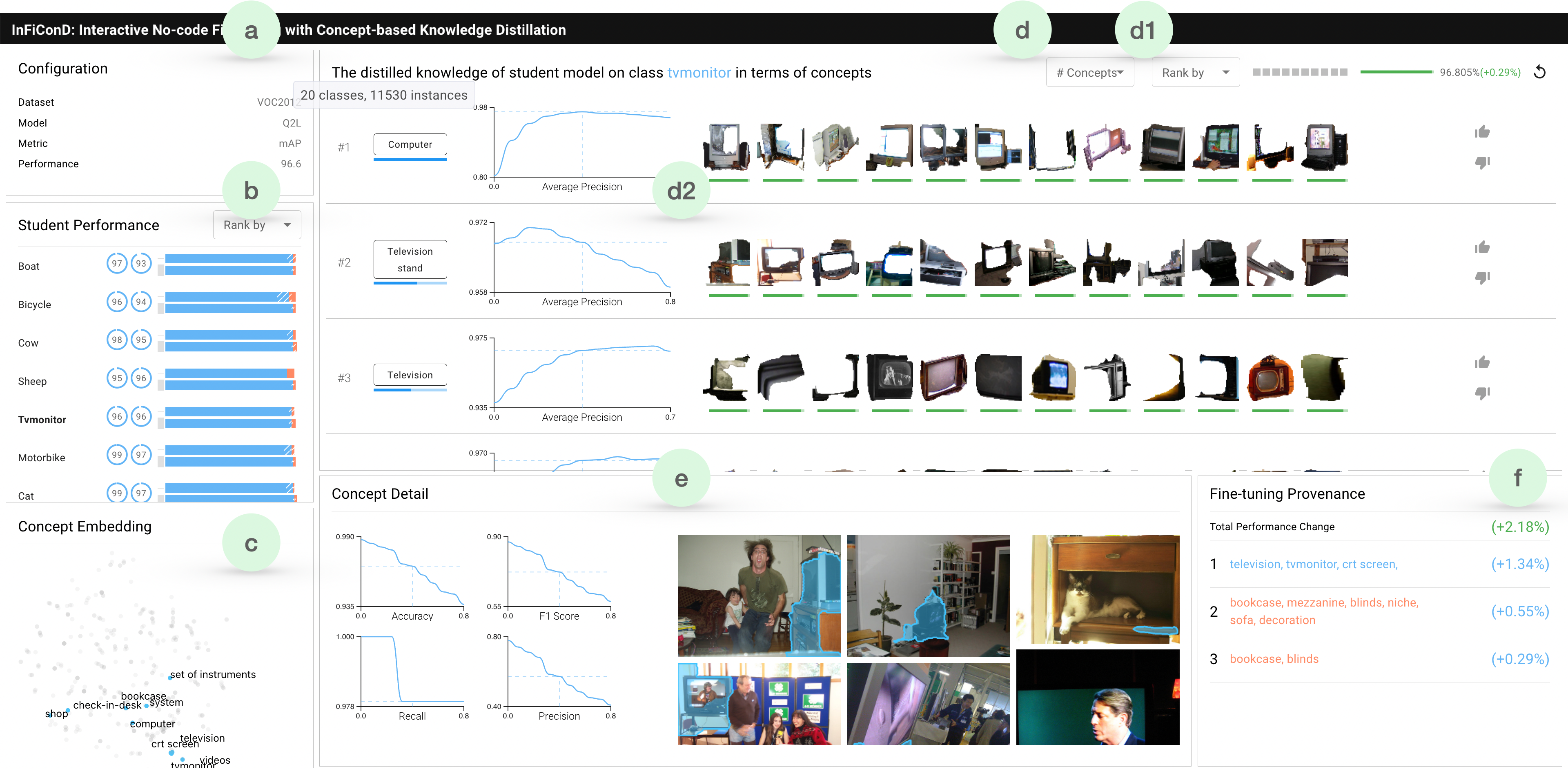}
  \caption{
  \system{}'s interface consists of six coordinated views: (a) The configuration view provides an overview of the dataset and teacher model being distilled, while (b) the student performance view displays a summary of each student model's performance and highlights subsets where student and teacher models misalign. (c) The concept embedding view illustrates the semantic relationships between influential concepts of a selected class. (d) The student knowledge view provides visualizes knowledge distilled into the student model, allowing users (d1) to efficiently identify crucial concepts of the student model (d2) gain concept-based performance insights and interactively tune concepts to improve student models. (e) The concept detail view provides additional performance insights and contextual details about a concept, enabling careful examination to prevent hasty conclusions. (f) The fine-tuning provenance view displays past concept tuning instructions and their impacts on performance.}
  \label{fig:teaser}
}

\graphicspath{{figs/}{figures/}{pictures/}{images/}{./}} 

\usepackage{tabu}                      
\usepackage{booktabs}                  
\usepackage{lipsum}                    
\usepackage{mwe}                       
\usepackage[dvipsnames,svgnames]{xcolor}
\usepackage{soul}
\usepackage{mathptmx}                  
\usepackage{wrapfig}
\usepackage{fontawesome}

\definecolor{blue500}{RGB}{33,150,243}
\definecolor{orange500}{RGB}{255, 87, 34}
\definecolor{red500}{RGB}{229, 57, 53}
\definecolor{green500}{RGB}{76, 175, 80}

\definecolor{lightgray}{gray}{0.9}
\definecolor{darkgray}{gray}{0.4}

\newcommand{\ci}[1]{\textcolor{blue500}{\concept{#1}}}
\newcommand{\cd}[1]{\textcolor{orange500}{\concept{#1}}}
\newcommand{\perfi}[1]{\textcolor{green500}{#1}}
\newcommand{\perfd}[1]{\textcolor{red500}{#1}}

\begin{document}


\maketitle
\section{Introduction}

Recent advancements in large pretrained models (PTMs) have significantly enhanced their performance across diverse computer vision tasks~\cite{han2021pre, wang2023large}. Developed using significant computational resources and trained on extensive, often proprietary datasets, these PTMs have developed robust visual representations that are flexible for a diverse set of tasks~\cite{long2022vision, lai2024empowering, wang2023large}. However, the large size and resource demands of PTMs often limit their practical application and deployment, 
particularly in resource-constrained environments~\cite{goldblum2024battle, liu2024lightweight, kamath2023deep}. Furthermore, the complexity of these models makes fine-tuning challenging, hindering their adoption in downstream applications~\cite{houlsby2019parameter, dinh2022lift}.

Knowledge distillation (KD) has emerged as a promising strategy for addressing these constraints by transferring knowledge from the large `teacher' model a more compact `student' model~\cite{gou2021knowledge, huang2022knowledge}. Importantly, KD has been shown as effective even when the teacher and student models exhibit significantly different architectures~\cite{lin2022knowledge, de2022structural, park2021learning}. In such cases, the process typically involves minimizing the difference in logits (inputs to the final softmax layer) produced by both models~\cite{hinton2015distilling, tung2019similarity, mirzadeh2020improved}. Despite the increasing adoption of KD as a means to overcome the limitations of PTMs~\cite{anil2018large, tian2019contrastive, jiao2019tinybert}, fine-tuning student models remains a non-trivial process, in part due to the low interpretability of student models and the lack of efficient fine-tuning mechanisms~\cite{alballa2024practical}.

In this paper, we investigate a novel strategy for KD based on interactively fine-tuning of student vision models in an interpretable and visualization-driven manner. In particular, we are inspired by recent efforts in KD interpretability that leverage \textit{visual concepts}—--a technique originally designed to explain model behaviors~\cite{kim2018interpretability, ghorbani2019towards, li2021visualizing}. Recent approaches have used concept-based approaches to reveal transferred knowledge \cite{cheng2020explaining,zhang2022quantifying, ge2021peek} or employ visual concept-based distillation algorithms for transferring knowledge to smaller models \cite{chen2022two, sousa2022conceptdistil}. 
While such methods can improve KD interpretability, they primarily rely on automated concept extraction pipelines that generate large ensembles of concepts in an unsupervised manner~\cite{ghorbani2019towards, sousa2022conceptdistil}, resulting in a significant percentage of ambiguous or irrelevant concepts~\cite{he2023clip}, and do not provide mechanisms to leverage these concepts to interactively fine-tune student models. In contrast, we investigate how enabling human users to interactively control the influence of concepts on student models can efficiently lead to their refinement and improvement.

Specifically, to address these challenges we design, develop, and validate a novel framework called \system{}, which supports interpretable and generalizable knowledge distillation from large pretrained vision models into compact linear student models, and enables efficient subsequent fine-tuning of the student model. \system{}'s backend (see Figure~\ref{fig:workflow}) leverages CLIP~\cite{clip} to extract and label visual concepts for downstream tasks such as multi-label classification~\cite{liu2021query2label}. Significantly, \system{} represents a \textit{no-code} solution for KD fine-tuning, which is increasingly recognized as a critical accessibility requirement across various domains where AI/ML models are being created, tuned, and deployed (including via KD~\cite{hinton2015distilling}) by domain users who are non-experts in the programming and/or the technical aspects of model development and distillation~\cite{cho2019efficacy, schizas2022tinyml, elton2024no}. \system{}'s frontend (see Figure~\ref{fig:teaser}) is a coordinated, multi-visualization interface to support fine-tuning workflows. This interface provides insights into the distilled student model’s knowledge (represented by influential visual concepts and their rankings), and supports interactive fine-tuning by weighting how concepts influence student models. 

The main contributions of this paper include the following:

(1)~Based on a conducted pre-study, we synthesize a set of current design challenges and goals for interpretable KD and subsequent interactive and no-code fine-tuning of student models.

(2)~To support these challenges and goals, we design and implement a unified framework in \system{}. \system{} incorporates an efficient backend pipeline for distilling knowledge using visual concepts and enables the subsequent fine-tuning of concept-based student models, paired with a novel visual interface that supports interactive fine-tuning activities in a no-code manner, including exploring underperforming subsets in the student model, examining their concept bases to identify causes of underperformance, and interactively manipulating the concept weights to improve the student model. 

(3) To successfully implement our pipeline, we develop several novel processes and algorithms, including: (i)~A model-agnostic method for distilling knowledge from PTMs into interpretable student models, ensuring generalizability. 
(ii) A pre-processing step that converts images into interpretable concept-based vectors, externalizing the PTM's knowledge. Linear student models are then trained using these vectors to mimic the teacher model's predictions, reducing capacity demand while enhancing the student's ability to replicate the teacher's predictions. The student model learns to assign robust weights to different concepts to complete tasks.
(iii)~A concept tuning algorithm that allows users to fine-tune the student model without coding. \system{} allows users to specify concepts that they want to uptune or downtune in the interface, adjusts them in the backend, and uses the adjusted weights to serve as new initializations to fine-tune the student model for a few epochs, effectively adapting the model based on user instructions.

(4) Based on our experience in creating and robustly evaluating \system{} (via conducting a multi-stage usage scenario and a rigorous user study), we discuss insights and lessons learned about how visualization-driven tools like \system{} can support KD workflows and activities, such as student model fine-tuning and optimization through interactive and interpretability-focused techniques

\section{Background and Related Work}
\label{sec:related_work}

This section provides a brief overview of knowledge distillation, and then discusses relevant related work at the intersection of visual analytics in knowledge distillation and visual concepts.

\subsection{Knowledge Distillation}
\label{sec:kd_background}

Knowledge distillation (KD)~\cite{gou2021knowledge} is the process of transferring knowledge from a large `teacher' PTM to a more compact `student' model. KD is particularly valuable when deploying a large model in a given environment that is infeasible due to computational or other resource constraints~\cite{gou2021knowledge}. As such, KD has emerged as a strategy for deploying compressed student models across a number of real-world devices and application areas, including autonomous driving, medical and healthcare, mobile devices, and smart homes~\cite{alkhulaifi2021knowledge, gou2021knowledge}. 
Broadly, KD methods can be categorized across three facets: the types of knowledge that are transferred, the strategy for distilling such knowledge, and the architectures of teacher and student models.

(i) The major \textit{types of knowledge} transferred during KD are response-based, feature-based, and relation-based. Response-based knowledge consists of the logits or prediction results from the teacher model, and the distillation involves training the student model to mimic the teacher's outputs~\cite{chen2017learning, hinton2015distilling, lee2018self}. Feature-based knowledge requires the student model's feature map to approximate that of the teacher ~\cite{Romero2014FitNetsHF, zagoruyko2016paying, kim2018paraphrasing}. Relation-based knowledge aims to align the similarity between pairs of feature maps in the student model with those in the teacher model \cite{yim2017gift, lee2018self}. While feature-based and relation-based knowledge have generally been shown to result in better performing student models, their reliance on access to teacher's internal information makes them less broadly applicable \cite{passalis2020heterogeneous, jin2019knowledge}. In contrast, response-based knowledge is model-agnostic and thus more broadly applicable \cite{gou2021knowledge}.

(ii) The three primary \textit{distillation strategies} are offline, online, and self-distillation. Offline distillation uses a static, pre-trained teacher model to guide the student~\cite{hinton2015distilling, Romero2014FitNetsHF}, while online distillation involves training the teacher concurrently with any student models \cite{mirzadeh2020improved, zhang2018deep}. Self-distillation is a specific type of online distillation that uses the same architecture for both the teacher and the student~\cite{zhang2019your, hou2019learning, mobahi2020self, phuong2019distillation}.

(iii) The design of the \textit{teacher-student architecture} presents significant challenges due to the ``capacity gap''---the inherent limitations of a simpler model’s ability to approximate the complex mappings of a more advanced model~\cite{gou2021knowledge}.  Careful design is essential to effectively enhance the student model’s capacity. For example, recent findings suggest that providing a more graduated learning process or a pre-prosessing step as teacher assistant for the student model can help bridge this gap~\cite{mirzadeh2020improved}.

\system{} supports response-based knowledge transfer, as we train a student model to mimic the teacher model's logits. This approach avoids reliance on teacher model information and makes the method more generalizable. To improve the student model's interpretability and facilitate fine-tuning, we design the student as a one-layer linear model--- see Section~\ref{subsect:concept_mapping} for a discussion about the advantages and constraints of this strategy. To address the evident capacity gap, we devise a pre-processing step to map images into vectors in a space spanned by high-dimensional interpretable vectors corresponding to visual concepts. This allows the student model to learn how to predict logits from these concept-based vectors, effectively reducing the task complexity. The details of this process are discussed in Section~\ref{sect:method}.

\subsection{Visual Concept-based Knowledge Distillation}
\label{sec:vis_concepts_background}

Concept-based methods have emerged as an effective tool to enhance the explainability of deep neural networks~\cite{hoque2022visual}. Originally proposed in TCAV (Testing with Concept Activation Vectors)~\cite{kim2018interpretability}, these methods aim to quantify the influence of human-understandable concepts on a neural network's prediction. For example, TCAV can measure the impact of the "stripe" concept on the classification of an image as a "zebra," providing intuitive explanations that align with human concepts. Such measurements can scale from a single instance to an entire class of instances and even to the entirety of a dataset. By offering explanations that are easily comprehensible and scalable across various data granularities\cite{kim2018interpretability, huang2022conceptexplainer, park2021neurocartography, bau2017network}, concept-based methods address the limitations of traditional perturbation-based approaches, such as lack of scalability and reliability~\cite{fong2017interpretable, selvaraju2017grad, simonyan2013deep, zhou2016learning, kindermans2019reliability}. 

Recent advancements have further showcased the potential of visual concepts in enhancing the explainability of KD methods~\cite{cheng2020explaining, zhang2022quantifying} and even enabling direct KD~\cite{gu2020introspective}. These approaches leverage visual concepts as the viechcle of transferring knowledge. By training a student model to assign weights to different concepts in a manner similar to the teacher model, the student effectively acquires the crucial knowledge necessary for the task at hand~\cite{li2020survey}. However, employing traditional TCAV-based concepts for KD presents three significant challenges:

(1) When dealing with a task that contains 
a vast concept space, manually annotating all relevant concepts becomes an arduous task. Although automatic concept extraction methods can alleviate this burden, they often fail to provide associated labels for the extracted concepts, hindering interpretability and offering no control over the concepts being extracted. These methods also have no capability of leveraging existing concept corpora to guide the extraction process, which further limits their usefulness.   (2) TCAV-based methods are computationally intensive and require architectural knowledge of the teacher model, which limits their applicability. (3) The absence of straightforward techniques to fine-tune the student model post-KD hinders the realization of the full potential of concept-based methods, including enabling intuitive understanding and interaction.

In \system{}, we overcome these challenges as follows: (1) Instead of relying on TCAV-based methods to generate concepts, we uses a well-defined knowledge corpus~\cite{he2023clip} as our concept list, paired with a CLIP-based method to extract text-labeled visual concepts according to this list. This approach ensures that the extracted concepts are interpretable and aligned with the desired domain knowledge for the task where we want to perform knowledge distillation (KD).  (2) Unlike existing work~\cite{gu2020introspective} that often uses feature-based knowledge distillation (KD) requiring internal knowledge about the teacher model, we utilize response-based KD to train the student model to assess concept weights. This method avoids dependency on teacher model knowledge, making our approach more generalizable and applicable to a wider range of models. (3) Significantly, we introduce a novel concept tuning method that allows users to fine-tune the student model by specifying concepts that need to be uptuned (increasing influences) or downtuned (decreasing influences). This technique enables users to fine-tune the student model to their specific needs without the burden of laborious manual programming. To the best of our knowledge, \system{} is the first system to achieve this level of user-friendly, interactive fine-tuning of concept-based models, empowering users to leverage the power of interpretable KD in their domains.

\subsection{Visual Analytics with Interpretable Concepts}
\label{sec:vis_analytics_int_concepts}

There is a growing number of visual analytics applications that utilize interpretable visual concepts for deep learning-related tasks, including data labeling, exploration~\cite{hoque2022visual, wang2022drava, zhao2021human}, and model explanation~\cite{huang2022conceptexplainer, park2021neurocartography}.

For data generation, ConceptExtract~\cite{zhao2021human} and Visual Concept Programming~\cite{hoque2022visual} propose interactive approaches to improve concept extraction and scalable image labeling using automatically extracted concepts. For model understanding, Drava~\cite{wang2022drava} takes a data-centric approach, generating interpretable concepts and enabling concept-based data exploration. NeuroCartography~\cite{park2021neurocartography} and ConceptExplainer~\cite{huang2022conceptexplainer} adopt a model-centric approach, with the former summarizing and visualizing learned concepts using neuron clustering and embedding, and the latter enabling structured exploration of a deep learning (DL) model's concept space and providing insights into model behavior across various data granularities.

However, these works focus primarily on using visual concepts for model explanation, not addressing the specific challenges of interpretability or fine-tuning in knowledge distillation (KD). In contrast, our work presents a comprehensive, closed-loop workflow that supports explainable KD of pretrained models (PTMs) through visual concepts and enables no-code model fine-tuning post-distillation, filling an important gap in the existing literature.
\begin{figure*}[t]
    \centering
    \includegraphics[width=.9\textwidth]{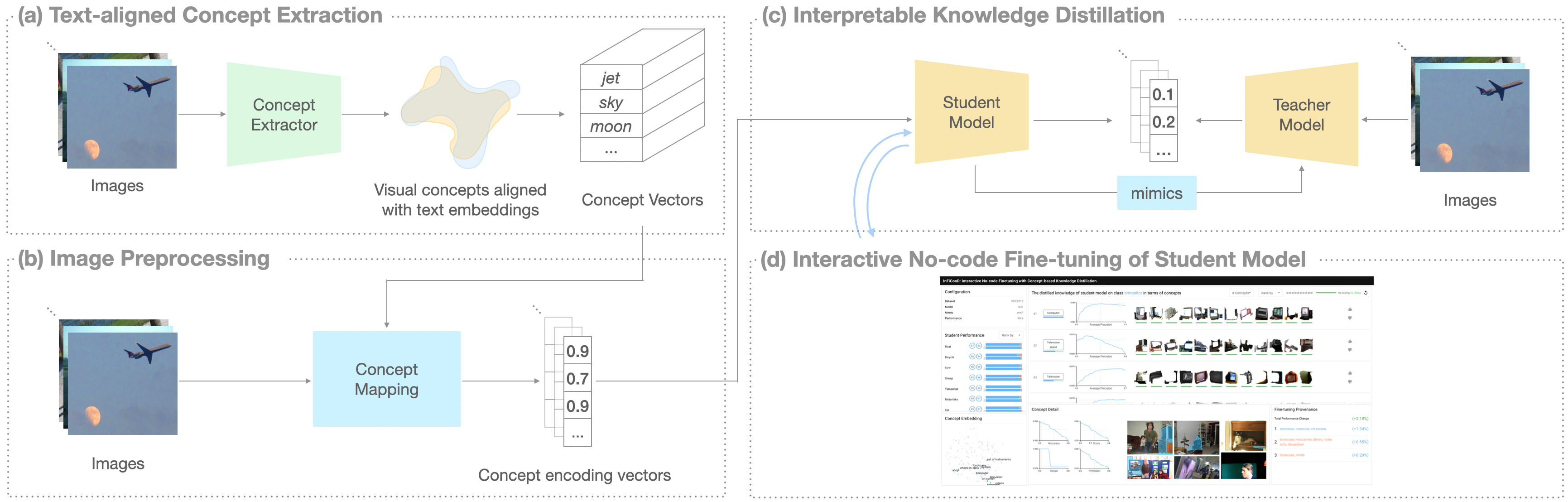}
    \caption{\system{}'s pipeline (a) extracts text-aligned visual concepts, (b) maps images into concept-based interpretable vectors, (c) trains linear student models that based on these vectors guided by teacher logits, and (d) uses a visual analytics interface to visualize  these concepts for the student model, emphasizing those with high influences to facilitate model analysis and fine-tuning.}
    \label{fig:workflow}
    \vspace{-.5cm}
\end{figure*}

\section{Formalizing Design Constraints for \system{}}
\label{sect:design}

\subsection{Design Challenges}

Our goal is to develop a framework that efficiently supports interpretable KD from parent models and subsequent (i.e., post-distillation) fine-tuning of student models, in a way optimizes human effort and accessibility (i.e., AI/ML non-experts can perform it). To help constrain our efforts, we reviewed prior interactive, no-code, and visualization-driven KD design studies (see Section~\ref{sec:related_work} papers), identifying four key design challenges for such a framework.


Existing methods (e.g., \cite{Romero2014FitNetsHF, zagoruyko2016paying, kim2018paraphrasing, passalis2020heterogeneous}) often prioritize student model performance at the cost of introducing complex architectures. This increased complexity hinders the adoption of KD\cite{gou2021knowledge} and negatively impacts subsequent fine-tuning processes, such as identifying the causes of underperformance, devising improvement strategies, and executing them effectively~\cite{wang2021knowledge, zhang2022quantifying}. Conversely, ``simpler'' student models can increase interpretability but also widen the capacity gap between the student and teacher models, making it challenging to transfer the desired knowledge effectively \cite{muller2019does}. Therefore, striking an appropriate balance between high student effectiveness and suitable interpretability is crucial for successful knowledge distillation and fine-tuning processes.

\textbf{(C2) Ensuring Generalizability in Knowledge Distillation.} While using more information from the teacher model can lead to better student model performance, it can also make the KD method heavily dependent on the specific teacher model architecture \cite{huang2024knowledge, son2021densely}, hindering its generalizability. Model-agnostic methods have been suggested for making KD more widely applicable\cite{zhu2018knowledge, shin2024teacher, huang2022knowledge}, but they come with their own challenges. For example, a model-agnostic strategy means that only the input and output of the teacher model (and any information derived from them) can be used to train the student model. This constraint can significantly limit the effectiveness of KD~\cite{gou2021knowledge}, especially when there is an additional requirement for interpretability. Developing a system that is generalizable across a variety of pretrained models (PTMs) while still maintaining high effectiveness is a non-trivial challenge, requiring careful consideration of the trade-offs between generalizability and performance in the context of interpretable knowledge distillation.

\textbf{(C3) Supporting Efficient Analysis of the Student Model.} Once the KD process is complete, users must be able to efficiently analyze the student model to validate its readiness for deployment. This can involve identifying and tracing the root causes of the model's current underperformance (e.g., \cite{zhu2022teach}) and learning how to address these issues\cite{rong2023towards}. From a user experience perspective, designing an interface that effectively supports this process requires careful consideration of the insights that should be revealed to the user and the types of actions/interactions that should be supported. Given that the user might not be an expert in AI/ML, there must be a balance between information depth (or complexity) and overall usability to ensure that users can analyze the student model effectively. The interface should provide meaningful and actionable insights while maintaining a level of simplicity that allows non-expert users to navigate and interpret the information presented, enabling them to make informed decisions about the student model's performance and potential improvements.

\textbf{(C4) Supporting Interactive Fine-tuning of the Student Model.} Once users understand the student model, they may wish to fine-tune it. The common practice for fine-tuning student models is largely manual and technical, requiring custom programming code to modify the model (e.g., \cite{mirzadeh2020improved}). This workflow hinders wider adoption of knowledge distillation (KD), particularly for domain users who conceptually understand how a student model should perform but lack the technical expertise to implement it in code. It also leads to inefficiencies, such as time spent writing code and introducing bugs or unoptimized logic~\cite{dankar2023improving}, diminishing the impacts of previous steps. Our solution is an interactive, no-code fine-tuning approach within the framework's interface. However, designing such a solution is not straightforward. Similar to \textbf{(C3)}, the user experience and interface must balance efficiency, control, and usability, providing intuitive mechanisms for users to fine-tune the model without programming expertise while ensuring the process remains efficient and yields meaningful improvements.

\subsection{Design Goals}

To address the challenges outlined above, we identify six key design goals that are subsequently operationalized in \system{}. These goals support the development of an interpretable and generalizable KD framework that facilitates  interactive analysis and no-code fine-tuning of student models.

\textbf{(G1) Enhancing Interpretability through Linear Student Models.} To tackle the challenge of interpretable KD \textbf{(C1)} and support efficient analysis \textbf{(C3)}, we investigate constructing the student model using a generalizable and highly interpretable architecture, specifically single-layer linear models. Although this choice has the potential to limit the student model's performance, we address this risk through a preprocessing method discussed in \textbf{(G2)} and (inspired by~\cite{zhu2018knowledge}) an ensemble-based approach that employs multiple linear models, each responsible for a specific subtask (see Section~\ref{subsect:model_choice_and_background} for more details).

\textbf{(G2) Representing Images as Interpretable Concept Vectors.} To compensate for the potentially reduced capacity resulting from using single-layer architectures \textbf{(G1)}, we introduce a preprocessing step that maps images into interpretable vectors \textbf{(C1)}. These vectors encode the visual concepts present in the original image, effectively capturing its critical information. This preprocessing step allows the student model to focus on learning the mappings between concept vectors and logits, simplifying the complexity of the learning task and ensuring the feasibility of a simple, interpretable, and performant KD process.

\textbf{(G3) Extracting Text-labeled Concepts using Multimodal Models.} The interpretable concept vectors used in the preprocessing step~\textbf{(G2)} rely on a list of interpretable visual concept vectors. However, existing concept extraction methods (e.g., \cite{ghorbani2019towards}) lack text descriptions for the extracted concepts, making them unsuitable for our purpose. To address this, we employ a concept corpus~\cite{he2023clip} and CLIP-based extraction method~\cite{he2023clip} to align image patches with the relevant words in the corpus, creating concepts with clear, interpretable meaning \textbf{(C1)}.

\textbf{(G4) Ensuring Generalizability through Model-agnostic Training.} To address the challenge of generalizable KD \textbf{(C2)}, we train the student model to mimic the teacher model using a response-based method (see Section~\ref{sec:kd_background}). The student model learns to replicate the teacher's input-output behavior, removing any dependencies on teacher model-specific information \cite{afonin2021towards}. This model-agnostic training also allows the KD process to generalize to different teacher models.

\textbf{(G5) Facilitating Efficient Analysis through Visualizations.} To support efficient analysis of the student model \textbf{(C3)}, we focus on two key aspects using a visualization-driven approach. First, we highlight the most underperforming subsets, providing users with a clear starting point for their analysis. Second, we show how the student model's performance changes in response to modifications of its parameters, such as a concept's influence. These visualizations enable users to identify the changes required to improve the student model's performance, facilitating effective fine-tuning.

\textbf{(G6) Enabling Interactive Fine-tuning through Concept-based Manipulation.} To address the challenge of interactive fine-tuning \textbf{(C4)}, we develop a workflow to accept user-specified tuning instructions. Based on user selections and interactions, a backend method automatically fine-tunes the model based on the provided instructions and returns the tuning results to the user. By supporting concept-based manipulation of the student model, we facilitate effective and interactive fine-tuning in a no-code manner.
\section{Methods}
\label{sect:method}

Based on the design goals outlined in Section~\ref{sect:design}, we develop the \system{} framework (shown in Figure~\ref{fig:workflow}). The process begins with the backend generating a set of text-aligned visual concepts \textbf{(G3)}. Next, each image is mapped into a vector that quantifies the presence of the set of concepts in the image \textbf{(G2)}. \system{} then distills knowledge from a teacher model into linear student models by training the student models to predict logits based on the vectorized images \textbf{(G1, G4)}. A visual analytics interface supports interactive performance analysis of student models, providing the user with fine-tuning insights. The user can interact with the interface to execute their insights for fine-tuning \textbf{(G5)}. This process is supported by a concept tuning method that translates user inputs into training constraints and automatically updates the student model's parameters, enabling no-code fine-tuning \textbf{(G6)}. In this section, we describe the primary backend components and processes in \system{}; the frontend interface is discussed in Section~\ref{sect:interface}.

\subsection{Application Scenario and Teacher Model}
\label{subsect:model_choice_and_background}

We demonstrate the \system{} framework using multi-label classification as an example application scenario. Multi-label classification represents a challenging, complex, and widely applicable AI task with high-stakes real-world applications such as airport security inspection~\cite{vluymans2018multi}; it is also an active research area for knowledge distillation (KD)~\cite{yang2023multi}. In multi-label classification, images can have multiple labels, meaning an image can contain many different objects. Due to the imbalance in class prevalence, achieving high accuracy across all labels, especially the less common ones, can be non-trivial and often requires sophisticated models \cite{liu4585915knowledge}.

To demonstrate how \system{} supports KD for multi-label classification, we use the Q2L model~\cite{liu2021query2label} as the teacher model. Q2L achieves state-of-the-art performance across multiple multi-label classification benchmarks, ranking \#1 on NUS-WIDE~\cite{chua2009nus}, PASCALVOC 2007~\cite{pascal-voc-2007}, and PASCALVOC 2012~\cite{pascal-voc-2012}, and \#5 on MS-COCO~\cite{lin2014microsoft}. The teacher model consists of a backbone for spatial feature extraction and multiple concatenated transformer decoders~\cite{liu2021query2label} for query and pooling mechanisms. The architectural complexity of Q2L makes it challenging for resource-constrained scenarios \cite{liu2021query2label}; however, it is important to note that \system{} is teacher model agnostic.

\begin{figure}[h]
    \centering
    \includegraphics[width=0.75\linewidth]{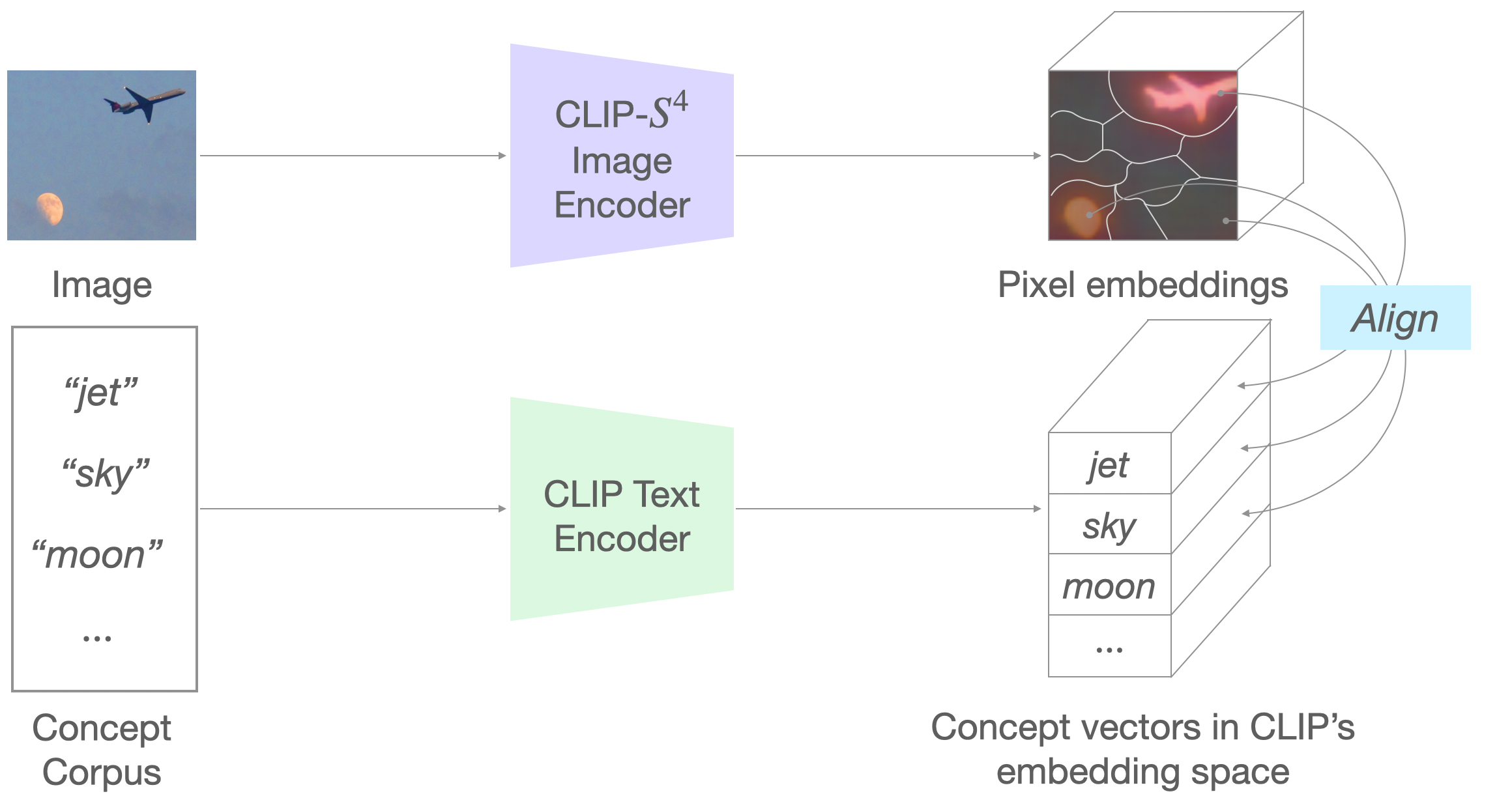}
    \caption{We employ the CLIP-S$^4$ model, trained to align pixel embeddings (image segments) with text embeddings, to extract text-labeled visual concepts.}
    \label{fig:conceptExtractor}
\end{figure}

\subsection{Text-Aligned Concept Extraction}

\system{} uses visual concepts (see Sections~\ref{sec:vis_concepts_background} and~\ref{sec:vis_analytics_int_concepts}) as a crucial element to achieve its design objectives. However, most existing solutions for automated concept generation, such as \cite{ghorbani2019towards}, primarily generate unsupervised pixel patches or embeddings, lacking human-interpretable labels or notations to describe these concepts. To address this limitation, we employ a novel process that extracts and quantifies text-aligned visual concepts from a given set of images.

The concept extraction process involves two main steps, as illustrated in Figure~\ref{fig:conceptExtractor}. Firstly, we use an image encoder from CLIP to segment images. Specifically, we utilize the CLIP-S$^4$ image encoder, which is fine-tuned from CLIP's image encoder to support semantically consistent segmentation \cite{he2023clip}. Pixels within the same segment exhibit similar embeddings, while those in different segments show significantly different embeddings.

Simultaneously, we encode a concept corpus using CLIP-S$^4$'s text encoder~\cite{he2023clip}. This corpus includes a 584-word list of concepts derived from sources such as PASCAL Context~\cite{mottaghi_cvpr14}, DAVIS 2017~\cite{perazzi2016benchmark}, COCO-Stuff~\cite{caesar2018coco}, and other concept works like \cite{bau2017network}. This results in 584 vectors, each representing a distinct concept.

For each image segment, we assign a label by identifying the concept vector that has the highest cosine similarity to it. Subsequently, for each concept text, we aggregate the best-matching segments. This process yields a list of text-aligned visual concepts, where each concept is represented by a pair consisting of (1) segment(s) and (2) the corresponding descriptive text.

\begin{figure}[h]
    \centering
    \includegraphics[width=\linewidth]{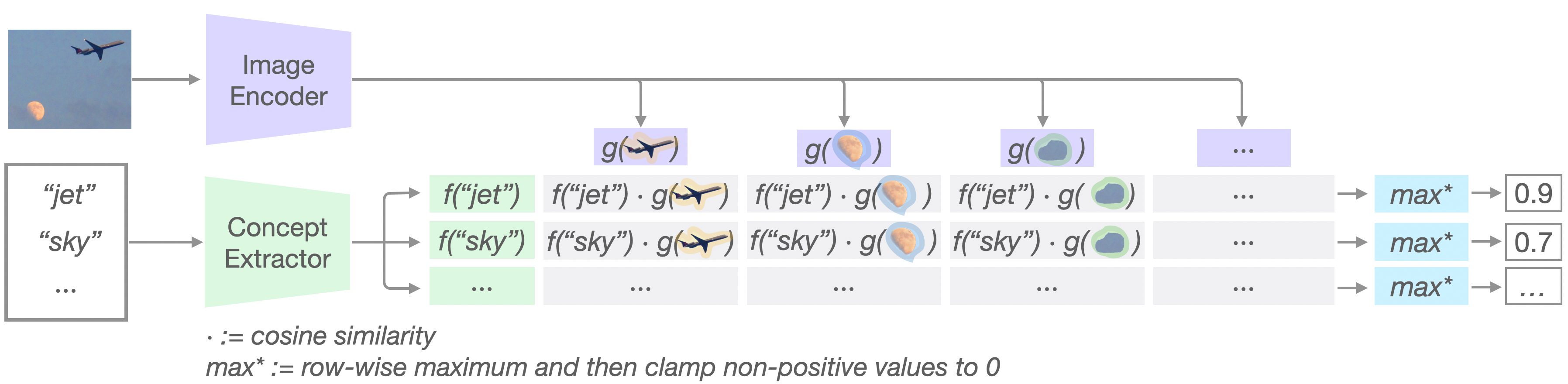}
    \caption{To perform concept mapping on an image, we compute the pairwise cosine similarity between its segment embeddings and concept vectors, and use the maximum cosine similarity value among all segments to represent a concept's degree of presence in the image. 
    }
    \label{fig:concept_presence}
\end{figure}

\subsection{Concept Mapping}
\label{subsect:concept_mapping}

As discussed for \textbf{(G1)}, linear student models have significant potential transparency and interpretability advantages. However, the simplicity of such an architecture might inhibit its capacity, limiting its potential performance relative to the teacher model. To address this risk, this step maps the explicit knowledge that needs to be distilled between the teacher and student models using the collection of previously extracted text-aligned concepts. By focusing the student model's responsibility on mapping the interpretable vectors to output logits, a ``simple but interpretable'' linear architecture becomes feasible (see Section~\ref{sec:discussion} for further discussion). This strategic shifting of the heavy lifting to a pre-processing step is a common practice in KD, known as introducing a teacher assistant~\cite{mirzadeh2020improved}.

In the context of multi-label image classification, the process involves mapping images to interpretable vectors, as illustrated in Figure~\ref{fig:concept_presence}. We leverage the segmentation and concept vectors derived during the previous concept extraction phase. For each image, we consider its segments and compute the cosine similarity between each concept and all segments from the image. The highest cosine similarity value represents the degree to which the concept is present in a part of the image. If the cosine similarity is less than 0, it indicates that no part of the image is similar to the concept, and the concept's presence in the image is considered to be 0. Conversely, if the cosine similarity is greater than 0, the concept is deemed to be present in the image to some extent, and the cosine similarity value is used to quantify this presence. Upon completion of the concept mapping process, each image is successfully transformed into a vector, where the $i$\textsuperscript{th} dimension of the vector represents the presence of the corresponding $i$\textsuperscript{th} concept in the image. Importantly, this activity only needs to be performed once as a pre-processing step, minimizing the overall computational overhead and supporting subsequent interactive workflows with the student models.

\subsection{Model-Agnostic Knowledge Distillation}
\label{subsect:model_agnostic_kd}

The next step is to distill knowledge from the teacher model using a response-based training scheme \cite{huang2022enhancing}. In this approach, only the teacher model's output guides the student model, enabling the student to predict consistently with the teacher model while also enhancing generalizability (G4). In the context of multi-label classification, the teacher model generates a multi-dimensional vector, where each dimension corresponds to a logit representing the likelihood of a particular class being present in an image. The teacher model produces a \textit{k}-dimensional vector, indicating the probabilities of \textit{k} classes (specifically in our application scenario, there are 20 possible classes available, so \textit{k=20}). To support a highly explainable student model, we utilize an ensemble of \textit{k} student models, with each model dedicated to predicting a single logit. This approach has previously been shown to be an effective strategy for being able to employ ``simple but effective'' student models  that do not enlarge the capacity gap~\cite{zhu2018knowledge}.

\begin{figure}[h]
\centering
\includegraphics[width=\linewidth]{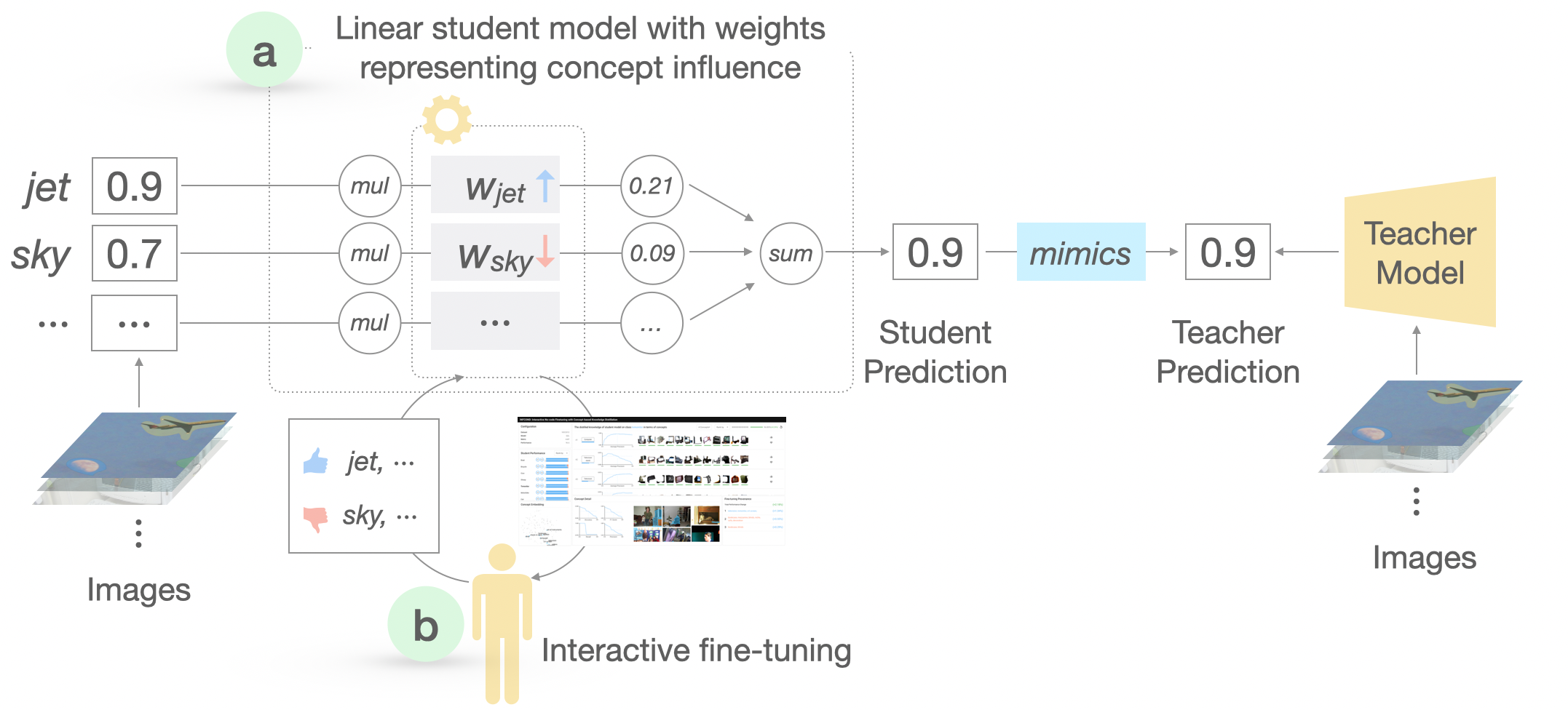}
\caption{(a) We train a linear student model for each class to imitate the teacher model's predictions. The student models consist of a single fully connected layer without activation functions. Each neural connection from an input node to the output node represents the influence of a concept on a specific class. (b)~Users can provide instructions specifying which concepts to tune and how to adjust their importance (i.e., increase or decrease).}
\label{fig:model_agnostic_distillation}
\end{figure}

The student models receive the 584-dimensional vectors as input~\ref{fig:model_agnostic_distillation}, where each vector represents an image characterized by the presence of 584 potential visual concepts. The student model's task is to map this 584-dimensional vector into a 1-dimensional logit. Consequently, the optimization process involves only 584 parameters, resulting in a highly interpretable and easily optimizable student model. In our case, the training process consists of 10 epochs and can be completed in approximately 30 seconds.

The weights of the $j$\textsuperscript{th} trained student model quantify the influence of each concept on the final prediction of class $j$. These weights are essentially the knowledge being transferred from the teacher model to the student model~\cite{yang2023categories}. Abstractly, we can view the student model as a learned assignment of weights to the concepts based on their presences to predict the occurrence of a particular class.

During the training process, we employ a binary cross-entropy loss function to measure the discrepancy between the teacher model's output logit and the student model's output logit. To promote sparsity and avoid dense weights where all concepts have small yet non-zero weights, we incorporate L1 regularization into the training process. The loss function is formally defined as follows:

\begin{equation}
    L = BCE(y_{student},y_{teacher}) + \lambda ||W_{student}||_{1}
\end{equation}

\noindent $BCE(\cdot,\cdot)$ is binary cross-entropy, $y_{student}$ and $y_{teacher}$ are student and teacher logits, and $||W_{student}||_{1}$ is L1 regularization on student weights, $\lambda$ is a hyperparameter to control how significant the sparsity penalty is, which is empirically demetrmined as $10^{-4}$.

\subsection{Fine-Tuning Student Models via Concept Tuning}
\label{subsect:concept_tuning}


The highly interpretable nature of the student models simplifies the analysis and fine-tuning process. Since the weights of the student models directly represent the influence of concepts on a particular class, fine-tuning the model to improve performance involves modifying these weights. To analyze a student model, users can determine which weights to adjust. However, due to the student model's convergence to a local minimum, merely increasing or decreasing a concept's weight as a hard constraint would not yield substantial changes.


To address this, we introduce a novel approach that utilizes the user's specification of concept uptuning or downtuning as soft constraints. \system{}'s interface allows users to review and select concepts that are deemed important for classification (e.g., uptuning \concept{flowerpot} for the \class{pottedplant} class) and concepts that are considered irrelevant (e.g., downtuning \concept{autobus} for the \class{dining table} class).

Based on user interactions (described in Section~\ref{subsect:student_knowledge_view}), we update the student models accordingly. For concepts that users identify as important, we establish a lower bound threshold for their weights and ensure that the weights never fall below this threshold during the training process. Conversely, for concepts that are considered irrelevant, we set an upper bound and guarantee that the weights never exceed this threshold. The lower and upper bounds are determined by scaling the original weights by a factor $x$. Users can provide instructions and fine-tune the students iteratively, maintaining a history of prior instructions. This instruction history is incorporated into the fine-tuning process to prevent the weights of previously selected concepts from surpassing the established thresholds. The fine-tuning process follows the same setup as the initial model training but with a reduced number of epochs to optimize computational resources. Upon completion of the fine-tuning process, the results are updated on \system{}'s frontend, allowing users to evaluate any performance improvements and provide further concept tuning instructions based on the updated results.
\section{Interface}
\label{sect:interface}

\system{}'s frontend (Figure~\ref{fig:teaser}) features six coordinated panels that enable interpretable knowledge distillation and interactive, no-code fine-tuning of student models. A demo video, included in supplemental materials, shows examples of the below described interface features and interactions.

\subsection{Configuration View}

The configuration view 
(Figure~\hyperref[fig:teaser]{\ref*{fig:teaser}a})
allows users to select and overview the dataset, the currently loaded student model, and performance metrics. Hovering over items reveals additional details on demand. For example, hovering over the dataset name displays the number of instances and classes within the dataset.

\subsection{Student Performance view}

The student performance view (Figure~\hyperref[fig:teaser]{\ref*{fig:teaser}b}) summarizes each student model's performance, emphasizing subsets where the student and teacher models disagree. Each student model corresponds to a specific class, sorted by its performance gap against the teacher model measured in average precision (AP) by default, though models can also be ranked by additional desired metrics including recall, precision, or F1-score. 

This view allows users to efficiently identify student models that require fine-tuning.
Each student model is labeled by its corresponding class and represented by two circular progress bars. The left bar indicates the teacher model's performance on that class, while the right bar represents the student model's performance. As an example, for class \textit{sofa} the student model has an AP of \perfd{60\%}, underperforming the teacher model by \perfd{17\%}, which has an AP of \perfi{77\%} (Figure~\ref{fig:usage_scenario_sofa})., while the bottom bar represents negative instances (images without a sofa). Blue areas indicate correct predictions by both models, and shaded blue showing student mispredictions. Orange areas represent mispredictions by both models, with shaded orange indicating correct predictions by the student and teacher mispredictions. The gray vertical bar's length next to each horizontal bar represents the relative size of the subset. A slim top bar indicates high dataset imbalance, with positive instances occurring at only one-tenth the ratio of negative instances.

These encodings provides detailed information about misaligned predictions between student and teacher models for two crucial reasons: (1)~it facilitates the efficient analysis of the KD process \textbf{(G5)}, and (2) it provides insights for subsequently fine-tuning student models \textbf{(G6)}. For example, a visible shaded blue area signifies that the student model mispredicts many cases where the teacher predicts correctly (a high false negative rate for the student). A user can tune the relevant influential concepts in the student knowledge view to address this issue and enhance the student's performance. The usage scenarios in Section~\ref{sect:usage_scenarios} show several examples of this process.

\subsection{Concept Embedding View}

The concept embedding view 
(Figure~\hyperref[fig:teaser]{\ref*{fig:teaser}c})
projects (via t-SNE) concept vectors from CLIP-S$^4$'s multimodal embedding space into a 2D space. This projection visualizes concepts and their semantic relationships, offering an overview of the concept landscape (i.e., the knowledge base) for the selected student model. When a user selects a student model, the view highlights the top 10 most influential concepts along with their text labels; other labels can be shown via hovering.

\subsection{Student Knowledge View}
\label{subsect:student_knowledge_view}

The student knowledge view  (Figure~\hyperref[fig:teaser]{\ref*{fig:teaser}d}) is shown when a student model is selected, and shows how knowledge is distilled into a student model in terms of its most influential concepts. This view provides users with performance insights and enables interactive concept tuning to improve the student model.

This view loads the top most influential concepts, sorted by their influence (by default, the top 10 concepts are shown, but this number can be adjusted via a dropdown in the submenu in 
Figure~\hyperref[fig:teaser]{\ref*{fig:teaser}(d1)}).
The submenu also supports sorting concepts based on their presence discrepancy in aligned vs. misaligned instances of positive or negative labels, helping identify concepts causing local mispredictions.

Each row in this panel corresponds to a specific concept (e.g., in 
Figure~\hyperref[fig:teaser]{\ref*{fig:teaser}(d2)}
the top row is \textit{Computer}). The leftmost number indicates the rank based on the current sorting method (weight, discrepancy (P), or discrepancy (N)). Clicking the concept name button for a row opens a concept detail view (described below) supporting detailed examination of the specific concept. The blue horizontal bar below the concept name button represents the concept's relative influence compared to the most influential concepts, highlighting potentially skewed concepts.

To the right of the concept name button, a line chart allows users to evaluate the potential effects that modifying a concept's influence will have on the student model's performance. The image patches to the right of the line chart provide visual representations of the concept, displaying the most similar patches to the concept vector in the embedding space. The green bars below each patche indicates its similarity.

Users can uptune (\faThumbsOUp) or downtune (\faThumbsODown) a concept using the corresponding buttons.  Clicking the uptune button turns the corresponding block in the top bar blue, while clicking the downtune button turns it orange.  After specifying the concepts to tune, users can see their options represented by colored blocks in the top bar, along with the current student performance and the overall performance change. Clicking the fine-tune button submits the instructions to the backend, which executes the concept tuning policy described in Section~\ref{subsect:concept_tuning} to fine-tune the student model. When the concept tuning process completes, the student knowledge view refreshes with an updated set of top influential concepts and student performance. Tuning instructions and their impacts are recorded in the fine-tuning provenance view (Figure~\ref{fig:teaser}f).

\subsection{Concept Detail View}
\label{subsect:concept_detail_view}
The concept detail view (Figure~\hyperref[fig:teaser]{\ref*{fig:teaser}e}) contains four line charts that visualize performance statistics for a selected concept's student model, and an image gallery showcasing examples of that concept.

The line charts display accuracy, F1 score, recall, and precision, providing insight into how the model's performance changes when concept influences are varied (thus helping users understand the concept's impact on the class). In particular, prior work has shown these metrics are effective for demonstrating model performance even for non-technical users \cite{tullio2007works}). For example, in Figure~\ref{fig:teaser}, all four metrics decrease as the television stand concept influence increases, indicating that further increasing the concept influence may not be beneficial. Additionally, the recall curve flattens once the television stand concept exceeds a value around \textbf{0.35}, suggesting that when the influence is too high, the model may start to miss more true positive instances (i.e., images containing a \class{TV monitor}).

Alongside the line charts, an image gallery supports qualitative review and confirmation of the concept (by showing example images containing it), enabling users to make well-informed decisions when uptuning or downtuning a concept.


\subsection{Fine-tuning Provenance View}

Finally, the fine-tuning provenance view  (Figure~\hyperref[fig:teaser]{\ref*{fig:teaser}f}) records previous concept tuning instructions and their impact on the student model's performance. Each row represents an entry, showing \textcolor{blue500}{uptuned} or \textcolor{orange500}{downtuned} concepts and the corresponding performance change from the previous entry (This color scheme is used consistently throughout the paper). The total performance change at the top-right corner displays the cumulative impacts of the fine-tuning.
\section{Usage Scenarios}
\label{sect:usage_scenarios}
To demonstrate the effectiveness of \system{}, we describe its usage by \user{}, who represents a non-technical user who wants to perform KD. Specifically, we demonstrate using the Q2L model trained on the PASCAL VOC 2012 dataset, which has 20 classes and contains 10,582 training and 1,449 validation images. Per the descriptions in Section~\ref{sect:method}, \system{} will use an extracted set of 584 text-aligned visual concepts, based on the concept corpus provided by~\cite{bau2017network}.

Each student model will have one fully-connected layer (no activation) of 584 input nodes and 1 output node. We use L1 regularization with weight $10^{-4}$ to balance student accuracy and its weight sparsity. We use a batch size of 2,084 and an Adam optimizer with a learning rate of 0.2. We also use batch normalization during training to stabilize the gradients. \user{} will train student models to mimic the Q2L predictions (i.e., using the Q2L model's outputs as ground truth). 

\subsection{Initial Performance Review}

\user{} begins by reviewing the performance of the student models and identifying key classes for improvement. Upon analysis, he discovers several classes in which the student model outperforms the teacher model, including \class{sheep}, \class{bird}, and \class{horse}. Specifically, for the \class{sheep} class, the student model achieves an AP of \perfi{96.25\%}, while the teacher model only achieves \perfi{95.31\%}. Similarly, for the \class{horse} class, the student model attains an AP of \perfi{98.18\%}, compared to the teacher model's \perfi{98.03\%}. For the \class{bird} class, the student model has an AP of \perfi{99.84\%}, surpassing the teacher model's \perfi{99.74\%}. Intrigued by these results, \user{} decides to fine-tune the \class{sheep} class to further improve performance. As a result, he boosts the \class{sheep} class by an additional \perfi{0.85\%}, bringing it to a total of \perfi{97.10\%}. Encouraged by this, he proceeds to examine other potential classes for improvement, particularly where the student model underperforms the teacher model.

\begin{figure}[h]
\centering
\includegraphics[width=\linewidth]{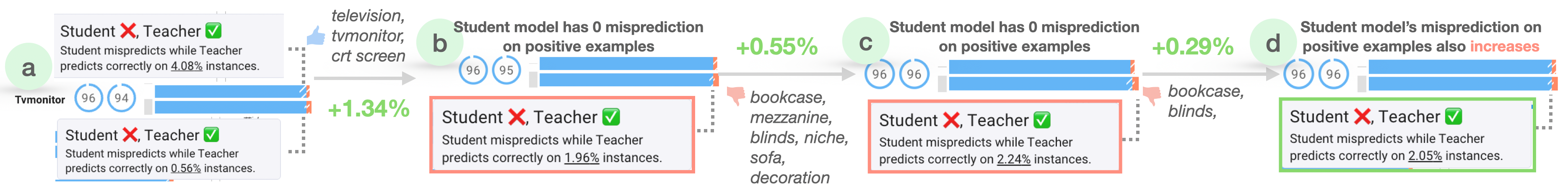}
\caption{In three iterations, \user{} improves the \class{tv monitor} class from underperforming by \perfd{1.41\%} to outperforming by \perfi{0.77\%}, with an overall average precision increase of \perfi{2.18\%}.}
\label{fig:usage_scenario_tv_monitor}
\end{figure}

\begin{figure*}[h]
\centering
\includegraphics[width=\linewidth]{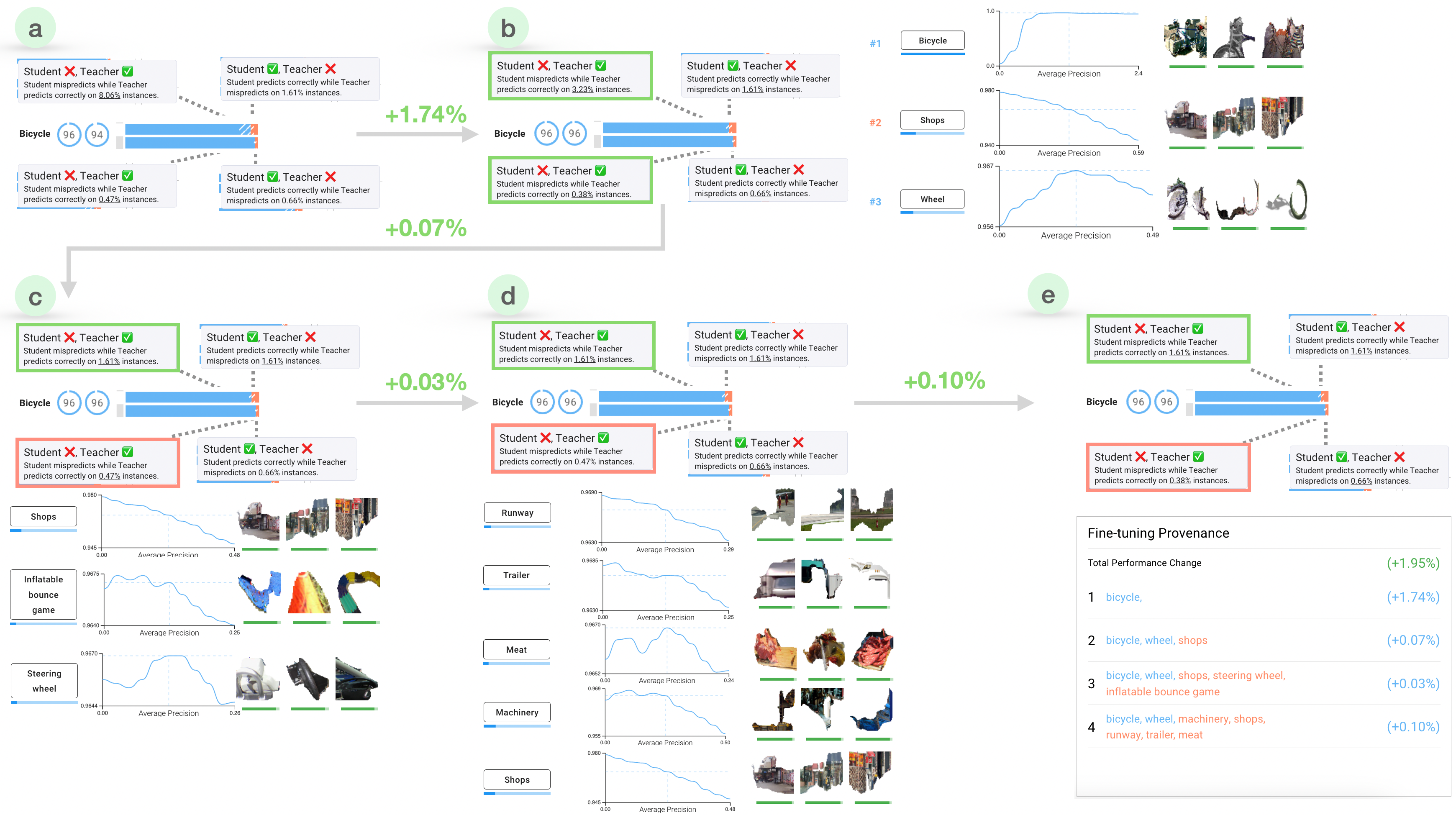}
\caption{In four iterations, \user{} successfully transforms the \class{bicycle} class from an underperforming class to an overperforming one. Starting by identifying the primary performance gap as the student model's inability to accurately predict positive samples, \user{} focuses on increasing the influence of concepts that are relevant to \class{bicycle}. He then examines the student knowledge view and reduces the influences of irrelevant concepts to decrease false negative rates. By taking a systematic and thorough approach, \user{} achieves significant improvements in the student model's performance by \perfi{1.95\%}.}
\label{fig:usage_scenario_bicycle}
\end{figure*}

\subsection{Improving Underperforming Classes}

\textbf{Improving the \class{tv monitor} class.} Using the class performance view, \user{} realizes for the \textit{tv monitor} class, the student lags behind the teacher models by \perfd{2\%}. He can see that the main performance gap occurs in the \class{tv monitor} class's shaded blue area 
(Figure~\hyperref[fig:usage_scenario_tv_monitor]{\ref*{fig:usage_scenario_tv_monitor}a}),
where the student mispredicts \perfd{4.08\%} of positive samples that are correctly identified by the teacher. To address this issue, \user{} needs to increase the student's accuracy at true positive samples while avoiding adding false positives. 

\user{} realizes this can be accomplished by increasing the influence of concepts highly relevant to \textit{tv monitors}. He clicks on the \ci{tv monitor} text to open the student knowledge view and updates the embedding view. He identifies concepts such as \ci{tv monitor}, \ci{crt screen}, and \ci{television} are reliable indicators and proceeds to examine the student knowledge view, sorting by weights. Though the \ci{tv monitor} and \ci{television} concepts already have high influences (ranked at \#2 and \#3, respectively), the performance charts suggest they can be further increased to achieve additional performance gains (Figure~\hyperref[fig:usage_scenario_tv_monitor]{\ref*{fig:usage_scenario_tv_monitor}b}).
\user{} uptunes these concepts and sees a significant improvement in performance. He scrolls down and notices that \concept{crt screen} only has an influence of 0.165, even lower than \concept{bookcase} and \concept{shop}. It clearly needs improvement, so he uptunes it as well. Satisfied with this tuning, \user{} submits his tuning instructions and gains a \perfi{1.34\%} improvement.

Upon rechecking the performance view, \user{} notes that while there is a \perfi{4.08\%} drop in false negative rate 
(Figure~\hyperref[fig:usage_scenario_tv_monitor]{\ref*{fig:usage_scenario_tv_monitor}b}), previously there were 4.08\% more mispredicted positive samples in student than in teacher, and now there are zero more. However, the false positive rate has also increased by \perfd{1.4\%}; previously there were only 0.56\% more mispredicted negative samples in student than in teacher, and now the gap increases to \perfd{1.96\%}. This means that, after the fine-tune process, some irrelevant concepts have gained increased influence which are causing some negative samples to be mispredicted as positive. To mitigate this unexpected confound and improve the student model's performance, \user{} decides to reduce the influence of irrelevant yet highly influential concepts. He goes through the list of concept and decreases the influence of irrelevant concepts, which include \cd{bookcase}, \cd{mezzanine}, \cd{blinds}, \cd{niche}, \cd{sofa} and \cd{decoration}. This gives him an additional improvement of \perfi{0.55\%}. He then goes through the list of concepts again and this time he sees only one irrelevant concept \cd{bookcase} that can be reduced to improve student performance. He does this and gains another \perfi{0.29\%} improvement.

Overall, \user{} has improved the performance of an underperforming class by \perfi{2.18\%}, resulting in a predictive performance that is \perfi{0.77\%} ahead of the teacher model. Despite not being technically trained in AI/ML, \user{} has interactively (and in a no-code manner) performed KD (while simultaneously improving the performance of student models) of the SOTA Q2L model while an AP that is already 96\%.

\textbf{Improving the \class{bicycle} class.} \user{} now addresses another significantly underperforming student model: \textit{bicycle}, which has a \perfd{1.89\%} AP gap between the student and teacher models (Figure~\hyperref[fig:usage_scenario_bicycle]{\ref*{fig:usage_scenario_bicycle}a})
In particular, the student model mispredicts \perfd{8.06\%} more positive samples compared to the teacher model, highlighting the need to increase the influence of highly relevant concepts.

\user{} examines concept relationships in the embedding view and identifies pertinent concepts such as \ci{bicycle} and \ci{wheel}.  In the student knowledge view (Figure~\hyperref[fig:usage_scenario_bicycle]{\ref*{fig:usage_scenario_bicycle}~(b)}), \user{} sorts concepts by presence discrepancy among correct and incorrect positive samples. He identifies the \ci{bicycle} concept as requiring enhancement, so he uptunes it. However, \user{} opts against uptuning the \ci{wheel} concept, as the performance chart does not indicate a performance increase by enhancing its influence (Figure~\hyperref[fig:usage_scenario_bicycle]{\ref*{fig:usage_scenario_bicycle}~(b)}). This decision is logical, as ``wheels'' are related, but not exclusive to, ``bicycles.'' \user{} checks the concept detail view for \ci{wheel} and confirms its presence in additional classes, including \class{motorbike} and \class{car}, further justifying the decision to maintain the \ci{wheel} concept's influence. By uptuning only the \ci{bicycle} concept, a \perfi{1.83\%} performance increase is achieved. 

Upon reexamining the performance view, \user{} observes a significant drop in false negative rate by \perfi{4.83\%} as the student model previously mispredicts \perfd{8.06\%} more positive samples than the teacher and it does only at \perfd{3.23\%} now (Figure~\hyperref[fig:usage_scenario_bicycle]{\ref*{fig:usage_scenario_bicycle}~(b)}). In the negative samples, the student model is also mispredicting fewer samples than the teacher model. It previously mispredicts \perfd{0.47\%} more and it is now mispredicting only \perfd{0.38\%} more. However, it used to have \perfi{0.66\%} more correctly predicted positive samples than the teacher model; it now only has \perfi{0.47\%} (Figure~\hyperref[fig:usage_scenario_bicycle]{\ref*{fig:usage_scenario_bicycle}~(b)}).

Examining the updated student knowledge view, \user{} notices that the \cd{shop} concept has more influence than the \ci{wheel} concept, which does not make intuitive sense. Thus, he tunes up \ci{wheel} and \ci{bicycle} while downtuning \cd{shop}. This brings him an increase of \perfi{0.07\%} (Figure~\hyperref[fig:usage_scenario_bicycle]{\ref*{fig:usage_scenario_bicycle}~(c)}). However, he notices that the false positive rate increases by \perfd{0.09\%}. To address this issue, he continues to downtune irrelevant concepts with unreasonable high influence and keeps uptuning \ci{bicycle} and \ci{wheel} to reinforce the effect. After downtuning \cd{shop}, \cd{steering wheel}, and \cd{inflatable bounce game}, he gains an improvement of \perfi{0.03\%}(Figure~\hyperref[fig:usage_scenario_bicycle]{\ref*{fig:usage_scenario_bicycle}~(d)}). In the subsequent iteration, after downtuning \cd{machinery}, \cd{shops}, \cd{runway}, \cd{trailer}, and \cd{meat}, he gains another \perfi{0.10\%} improvement (Figure~\hyperref[fig:usage_scenario_bicycle]{\ref*{fig:usage_scenario_bicycle}~(e)}). The student model finally reaches \perfi{96.82\%} of AP and starts to outperform the teacher model by \perfi{0.06\%}. Overall, \user{} is pleased with his outcome and moves on to address other underperforming classes.

\subsection{Mitigating Severe Underperformance}

\begin{figure}[h]
\centering
\includegraphics[width=\linewidth]{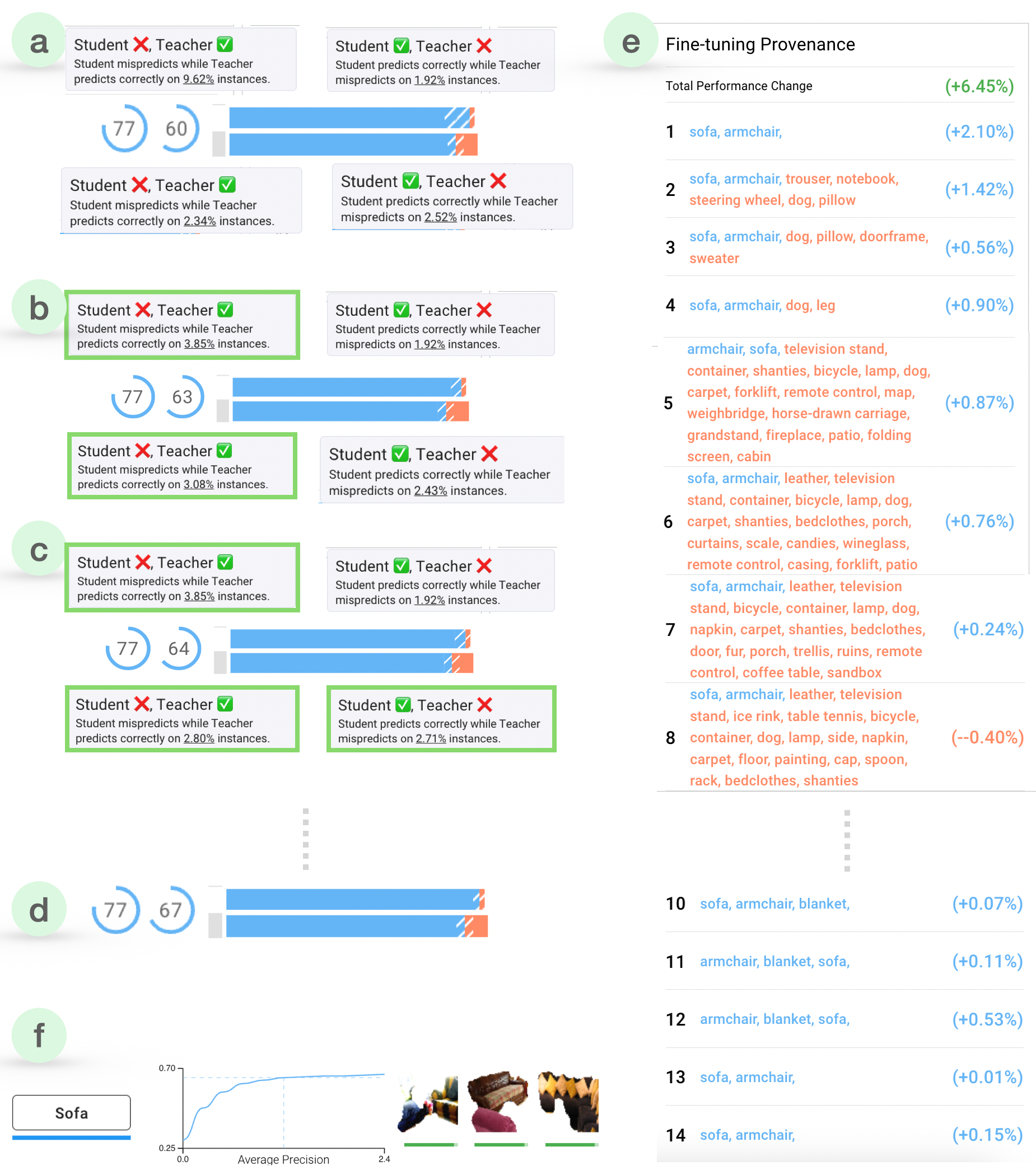}
\caption{An illustration for how \user{} mitigates severe underperformance of the \class{sofa} class}
\label{fig:usage_scenario_sofa}
\end{figure}

\textbf{Mitigating underperformance of the \textit{sofa} class.} After successfully improving two underperforming student models, \user{} sets out to fine-tune student models. He starts by addressing the severe underperformance of the \class{sofa} class, where the student model lags the teacher model by \perfd{17.0\%}. To address this, \user{} examines the performance view for the \class{sofa} class and observes a \perfd{9.62\%} more misprediction in positive samples by the student model than the teacher model (Figure~\hyperref[fig:usage_scenario_sofa]{\ref*{fig:usage_scenario_sofa}(a)}), which indicates the need to increase the influence of relevant concepts.

After scanning the student knowledge view, \user{} uptunes the \ci{sofa} and \ci{armchair} concepts (as armchairs are very similar to sofas). Only uptuning these two concepts results in a \perfi{2.10\%} improvement of the student model (Figure~\hyperref[fig:usage_scenario_sofa]{\ref*{fig:usage_scenario_sofa}~(b)}).

Via the performance view, \user{} sees a reduction of false negative rates by \perfi{5.77\%}, but also an increase in false positive rates by \perfi{0.74\%}. To address this, he decides to downtune irrelevant concepts and further uptune the \ci{sofa} and \ci{armchair} concepts. After reivew of available concepts, he downtunes \cd{trouser}, \cd{notebook}, \cd{steering wheel}, \cd{dog}, and \cd{pillow}, resulting in a \perfi{1.42\%} increase (Figure~\hyperref[fig:usage_scenario_sofa]{\ref*{fig:usage_scenario_sofa}~(c)}).

As \user{} iteratively tunes the student model, he gains incremental improvements each time. After 14 iterations (Figure~\hyperref[fig:usage_scenario_sofa]{\ref*{fig:usage_scenario_sofa}~(e)}), the student model reaches \perfi{67.0\%} of AP, which is 7\% higher than the initial 60\% of AP (Figure~\hyperref[fig:usage_scenario_sofa]{\ref*{fig:usage_scenario_sofa}~(d)}). The performance view shows a significant reduction in the shaded blue area, which indicates a large decrease in false negatives. The performance chart of the \class{sofa} class also suggests that if \user{} continues to uptune the \ci{sofa} concept, he can eventually reach 70\% of AP for the \class{sofa} class (Figure~\hyperref[fig:usage_scenario_sofa]{\ref*{fig:usage_scenario_sofa}~(f)}).

\textbf{Mitigating underperformance of the \textit{dog} class.} After successfully mitigating the performance gap in previous classes, \user{} turns his attention to improving the student model's performance in the \class{dog} class. The initial performance of the student model is quite low, with only \perfi{90.72\%} of AP compared to the teacher model's 99.94\%. Upon examining the performance view, \user{} identifies the primary issue as a high false negative rate, with the student model mispredicting \perfd{9.09\%} more positive samples than the teacher model.

To address this issue, \user{} decides to enhance the influence of the \ci{dog} concept, which is the most relevant and influential concept for the \class{dog} class. In the first iteration, uptuning the \ci{dog} concept alone brings an improvement of \perfi{2.02\%} and reduces the false negative rate by \perfi{1.17\%}. However, the student model still mispredicts \perfd{7.92\%} more positive samples than the teacher model, so \user{} continues uptuning the \ci{dog} concept. Besides uptuning the \ci{dog} concept, he also uptunes \ci{fur} and downtunes \cd{grass}, \cd{terraces} and \cd{apron} concepts, which are concepts that can co-occur with \class{dog}s but may lead to false positives.

In the second iteration, \user{}'s tuning results in an improvement of \perfi{0.48\%} AP and reduces the false negative rate to \perfi{5.94\%}. Encouraged, \user{} continues to uptune the \ci{dog} concept in subsequent iterations and downtuning other irrelevant yet influential concepts such as \cd{grass} and \cd{cushion}. After eight rounds, the student model reaches 94\% AP, an improvement of \perfi{3.28\%} from the initial performance, and mispredicts only \perfd{3.74\%} more positive samples than the teacher model. With this performance gap significantly reduced, \user{} decides to move on to other classes to further improve the student model.




\section{User Study}
\label{section:user_study}

To empirically validate \system{}, we conducted a user study with ten participants to answer our primary question related to design goals \textbf{(G1-G5)}: ``\textit{How well does the system support interpretable KD and subsequent fine-tuning?}'' The study design consisted of a system walk-through and two tasks, and we collect and analyze both quantitative and qualitative results, allowing us to robustly evaluate how \system{} supports  (G1--G5) while also assessing the system's overall usability.

\textbf{Participant and Study Setup.}
We recruited ten graduate computer science students from an anonymous university.\footnote{Anonymized for review} All participants self-reported as users of various PTMs, though none were experts in technical/programmatic KD workflows. Each participant took approximately 30 minutes (SD = 5 minutes) to complete the study. Participants viewed \system{} on a 30'' monitor with a resolution of $3840\times2160$ using Google Chrome in full-screen mode. The study sessions were conducted in a quiet, distraction-free office environment.

\textbf{Study Design.}
Participants completed a brief demographic questionnaire and received a brief introduction to the topic of multi-label classification and visual concepts. The administrator conducted a walk-through of the available features and functionalities of \system{}, and participants could opt to complete a simple training task to familiarize themselves with the system. 
During this training time, participants were free to ask questions at any time and explore the interface until they felt comfortable to proceed. After the study, users completed a post-study questionnaire to provide feedback to the system, and could provide additional freeform feedback if desired.

In the first task, participants were required to to understand and improve the \class{bicycle} class. The task was considered complete when the participant improved the performance of the student model to be higher than the teacher model. The average completion time for this task was 7 minutes (SD = 2 minutes). In the second task, participants were required to understand and improve the \textit{tv monitor} class via the same think aloud policy. The average completion time for this task was 4 minutes with a standard deviation of 1 minute. For both tasks, participants were encouraged to use think aloud during this task, especially to help explain their rationales when making decisions (e.g., uptuning a specific concept).

\begin{figure}[h]
    \centering
    \includegraphics[width=\linewidth]{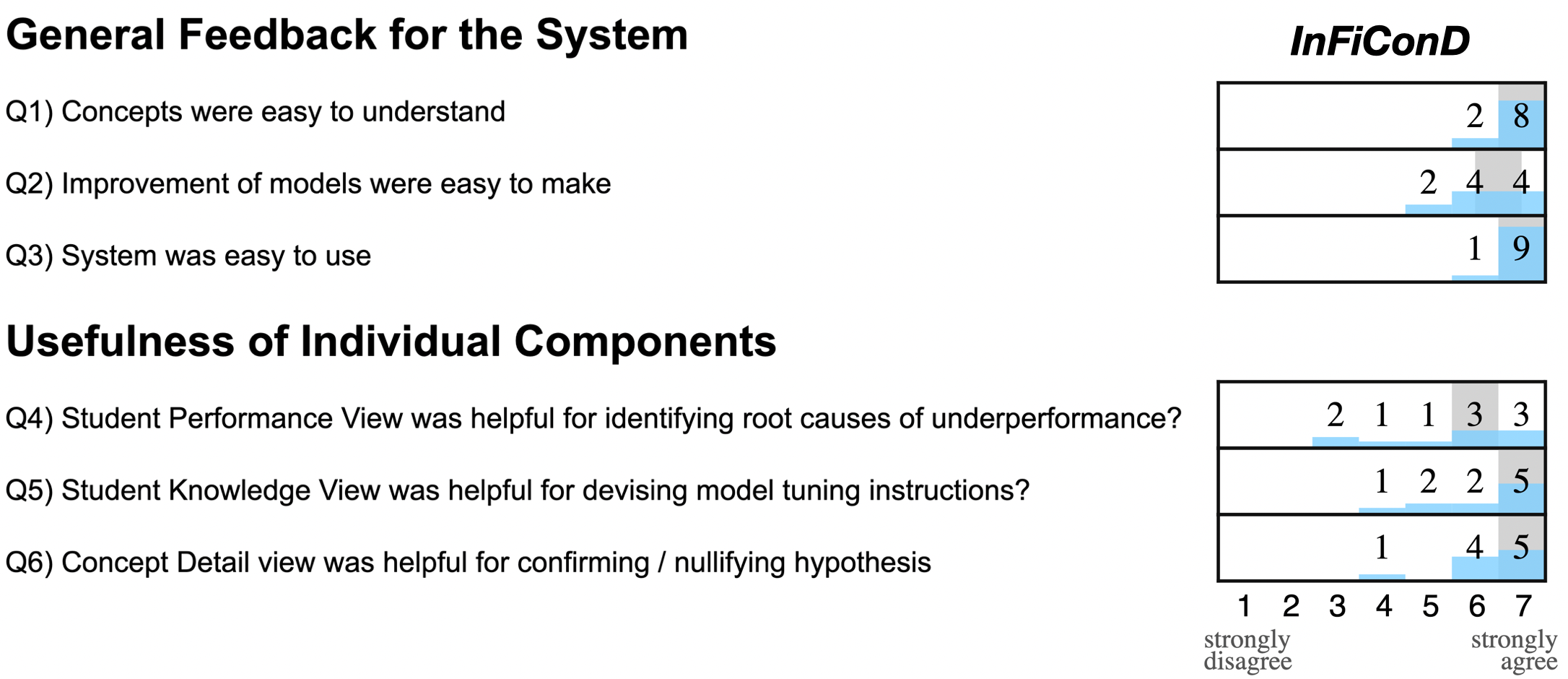}
\caption{Usability ratings for \system{} from the user study. The median ratings for each question are indicated in gray.}
    \label{fig:user_study}
\end{figure}

\subsection{Study Results}

In terms of KD performance and student model fine-tuning, all ten participants were able to improve the student model such that it outperformed the teacher model for both tasks. For the \textit{bicycle} task, students performed an average of 4.5 fine-tuning operations (SD = 0.5) resulting in an average performance increase of 1.91\% (SD = 0.04). For the \textit{tv monitor} task, the average was 2.5 fine-tuning operations (SD = 0.5) with a performance increase of 2.04\% (SD=0.05).

Figure~\ref{fig:user_study} shows responses for the post-study questionnaire. \textbf{Q1}--\textbf{Q3} ask about the overall system usability, and \textbf{Q4}--\textbf{Q6} focus on specific interface components.  Overvall, \system{} was highly rated. 

To analyze the think aloud verbalizations and post-study feedback, we performed an open coding process to understand the strengths and limitations of the \system{} user experience. We report several highlights below:


\textbf{Text-aligned visual concepts are easy to understand and work on.} All users understood the visual concepts immediately and were able to use the concept's semantic relationship to the class to come up with fine-tuning operations. No participants reported being confused with any concepts, which indicates that our text-aligned concept extraction and mapping methods successfully made the concepts understandable and removed ambiguity.

\textbf{The concept detail view provides necessary context to confirm understanding of visual concepts.} The participants heavily relied on the concept detail view during their tasks. Specifically, almost all participants (with the exception of P7) used this view to confirm their understanding of the concept. For instance, one participant (P3) stated, ``\cd{remote control} \textit{can be with a} \class{tv monitor}, \textit{but it can also be on its own. So I'm checking the details of} \ci{remote control} \textit{to see if they are mostly on a \class{tv monitor} or not.}'' Another (P5) mentioned, ``\textit{I'm not sure what} \cd{set of instruments} \textit{means here, so I need to look at the details.}''

\textbf{The student performance view illustrates where to start the analysis.} The majority of participants (8/10) began their analysis by examining the largest shaded areas in the student performance view and prioritizing concepts accordingly in the student knowledge view. For instance, one participant (P4) remarked, ``\textit{The shading feature is helpful for me to figure out where to start},'' while another (P9) stated, ``\textit{I'll fix the ones with the biggest shade first.}''

\textbf{The student knowledge view enables effective analysis and interaction.} Participants effectively interacted with the student knowledge view to analyze and adjust concepts. 
All participants reported that the leaderboard was helpful and efficient. Per (P3), ``\textit{It's cool that I can just vote for the concepts and the model will update.}" (P5): ``\textit{I like the voting thing. It's intuitive.}'' (P8): ``\textit{It provides a really good indication of effects of tweaking each feature at a certain step.}'' 
Five participants used the ranking feature to sort concepts based on the presence discrepancy, as a way to identify concepts to tweak. Other participants preferred to sort concepts by their weight. 
In contrast to the fine-tuning strategies of other participants, P7 preferred to make minimal changes for each adjustment, to accurately identify the cause of any changes made: ``\textit{I want to focus on a few concepts at a time so I can identify the root cause for each change}.'' 

\textbf{An accessible workflow and interface.} Both the survey responses and verbal feedback indicate that our overall workflow, and in particular the ability to support no-code fine-tuning, supported accessibility and ease of use. (P8): ``\textit{Simple to interact. Easy to refine the model on a single click.}" (P2): ``\textit{The front-end design was good. Easy to understand.}'' (P10): ``\textit{Easy to train (fine-tune) and select features (concepts).}''

\section{Discussion and Future Work}
\label{sec:discussion}

\textbf{Extending \system{} to diverse computer vision tasks.}  Although we demonstrate \system{}'s effectiveness in multi-label classification, its potential extends beyond this specific application scenario. Rapid advancements in large PTMs have accelerated KD activities across a wide range of computer vision tasks, such as object detection~\cite{zou2023object} and semantic/instance segmentation \cite{minaee2021image}. To fully realize this potential, future work will focus on training the student model to incorporate concepts not only for classification but also for segmentation tasks. By doing so, we aim to expand the applicability of \system{} to enable efficient KD and fine-tuning in boarder tasks.

\textbf{Supporting user-specified concept corpora.} \system{} uses visual concepts as its ``vehicle of knowledge,'' and these are generated using an established (and pre-defined) concept corpus~\cite{he2023clip}. While this corpus provides a solid and validated foundation, it may face limitations when applied to specialized domains. For example, performing KD for medical tasks (e.g., tumor detection~\cite{abdusalomov2023brain}) might require an updated or customized concept corpus containing domain-specific concepts. \system{} is ``concept corpus agnostic,'' meaning a newer or different corpus could be swapped in, but the system could also be extended to allow expanding an existing corpus with additional concepts. A significant benefit to this approach is that such modifications \textit{do not} require the CLIP-S$^4$ encoders to be retrained. By supporting such flexibility, frameworks like \system{} can be generalized across a range of domain-specific applications.

\textbf{Promoting fairness through concept-based knowledge distillation.} The automatic nature of KD methods can lead to biases being transferred to student model as knowledge, if fairness control mechanisms are not implemented during the distillation process~\cite{zhou2022towards, madaio2020co}. Tools like \system{} can mitigate this issue by leveraging a concept-based approach, employing a carefully curated, de-biased concept corpus with additional human-in-the-loop oversight. By ensuring that the student model learns to perform tasks using unbiased concepts, we can promote the development of fair and unbiased models in KD.

\textbf{Facilitating language model fine-tuning with \system{}-inspired designs.} The remarkable capabilities demonstrated by large language models (LLMs) across various NLP tasks have lead to their applications in almost everywhere. However, the training process for LLMs typically involves a general training step that relies on data crawled from the entire internet~\cite{ouyang2022training}; a task that is prohibitively expensive for individual developers due to the immense computational and infrastructure requirements. Subsequent fine-tuning steps, which involve supplying high-quality, task-specific examples to teach the model to answer questions accurately, are more accessible but still require coding expertise. As more foundational models become open-sourced, there is a growing opportunity to democratize the fine-tuning process. By drawing inspiration from \system{}'s design principles (i.e., extending them to NLP or LLM doamins), we envision the development of no-code fine-tuning tools that empower developers to adapt open-sourced foundational models to create personalized LLM-based assistants. This direction holds promise for making the power of LLMs accessible to a wider audience.


\section{Conclusion}

We present \system{}, a novel framework that leverages visual concepts created by multi-modal models to implement interpretable knowledge distillation of large pretrained models and enable subsequent interactive fine-tuning of student models. We demonstrated the effectiveness of \system{} on a state-of-the-art multi-label classification model, where \system{} empowers users to analyze the student models in terms of influential visual concepts and fine-tune them interactively by specifying concept tuning instructions and running a concept tuning method to adjust concept influences. Our concept-based knowledge distillation approach represents a significant step towards developing efficient and user-friendly knowledge distillation systems that facilitate the application of knowledge distillation to a broader user base.

\bibliographystyle{abbrv-doi-hyperref}

\bibliography{template}

\end{document}